\documentclass[letterpaper, 10pt, conference]{ieeeconf}

\IEEEoverridecommandlockouts
\usepackage[T1]{fontenc}        
\usepackage{bm}                 
\usepackage{graphicx}
\usepackage{amsmath}
\usepackage{amssymb}
\usepackage{cite}
\usepackage{url}
\usepackage{booktabs}
\usepackage{multirow}
\usepackage{pifont}
\usepackage{array}
\usepackage{stfloats}

\newcount\irospagelimit \irospagelimit=8
\AtEndDocument{%
  \ifnum\value{page}>\irospagelimit
    \GenericWarning{}{IROS FORMAT WARNING: Paper is \the\value{page} pages,^^J%
      exceeding the IROS 2026 limit of \the\irospagelimit\space pages!^^J%
      You MUST reduce to \the\irospagelimit\space pages or fewer before submission.}%
  \fi
}

\usepackage{censor}
\StopCensoring  
\newif\ifcameraready
\camerareadytrue  
\usepackage{microtype}          
\makeatletter
\let\NAT@parse\undefined        
\makeatother
\usepackage[hidelinks]{hyperref} 

\title{\LARGE \bf AsyncMDE: Real-Time Monocular Depth Estimation\\
via Asynchronous Spatial Memory}

\ifcameraready
  \author{Lianjie Ma$^{1}$, Yuquan Li$^{1}$, Bingzheng Jiang$^{1}$, Ziming Zhong$^{3}$, Han Ding$^{2}$, and Lijun Zhu$^{1,\dagger}$%
  \thanks{This work was supported by the Fundamental and Interdisciplinary Disciplines Breakthrough Plan of the Ministry of Education of China under Grant JYB2025XDXM208, and the National Natural Science Foundation of China under Grant U25A6013.}%
  \thanks{$^{\dagger}$Corresponding author.}%
  \thanks{$^{1}$School of Artificial Intelligence and Automation, Huazhong University of Science and Technology, Wuhan, China. {\ttfamily\scriptsize \{yingyi1048596, yuquanli, bzjiang, ljzhu\}@hust.edu.cn}}%
  \thanks{$^{2}$School of Mechanical Science and Engineering, Huazhong University of Science and Technology, Wuhan, China. {\ttfamily\scriptsize dinghan@hust.edu.cn}}%
  \thanks{$^{3}$Yichang Testing Technology Research Institute, Yichang, China. {\ttfamily\scriptsize zhongziminghaha@outlook.com}}%
  }
\else
  \author{Author Names Omitted for Double-Blind Review}
\fi

\begin{document}

\bstctlcite{IEEEexample:BSTcontrol}

\maketitle
\thispagestyle{empty}

\begin{abstract}
Foundation-model-based monocular depth estimation offers a viable alternative to active sensors for robot perception, yet its computational cost often prohibits deployment on edge platforms. Existing methods perform independent per-frame inference, wasting the substantial computational redundancy between adjacent viewpoints in continuous robot operation. This paper presents AsyncMDE, an asynchronous depth perception system consisting of a frozen foundation model and a lightweight fast path that amortizes the foundation model's computational cost over time. The foundation model periodically produces high-quality spatial features in the background, while the lightweight fast path runs asynchronously in the foreground, fusing cached memory with current observations through complementary fusion, outputting depth estimates, and autoregressively updating memory. This enables cross-frame feature reuse with bounded accuracy degradation. With 3.83M trainable fast-path parameters and a 97.5M frozen slow path, AsyncMDE's fast path operates at 237~FPS on an RTX~4090, recovering 77\% of the accuracy gap to the foundation model. Across indoor static, dynamic, and synthetic extreme-motion benchmarks, AsyncMDE degrades predictably and reaches 161~FPS fast-path inference on a TensorRT-optimized Jetson AGX Orin, supporting real-time edge deployment.
\end{abstract}

\section{INTRODUCTION}\label{sec:intro}

Depth perception is a fundamental capability for embodied intelligent systems, playing a key role in autonomous navigation~\cite{yang_iplanner_2023, roth_viplanner_2024, shah_vint_2023, sridhar_nomad_2024}, perceptual control~\cite{han_dual_2024, nvidia_gr00t_2025, cui_openhelix_2025}, and vision-language-action decision making~\cite{kim_openvla_2025}. Mainstream approaches use LiDAR or RGB-D cameras, but these active sensors are costly and constrained by lighting conditions, sensing range, and environmental structure, making it difficult to maintain stable performance across diverse deployment scenarios. Monocular depth estimation (MDE) requires only a single RGB camera to produce dense, structured depth maps, offering low cost, minimal calibration, and broad adaptability across environments, which makes it an attractive alternative for robot perception~\cite{xu_towards_2026}. Recently, depth foundation models~\cite{yang_depth_2024, hu_metric3d_2024, piccinelli_unidepthv2_2026, bochkovskiy_depth_2024} have demonstrated strong zero-shot generalization through large-scale pretraining and Vision Transformer (ViT) architectures. However, their large parameter counts lead to excessive inference latency on edge platforms, failing to meet the real-time perception demands of robots operating in highly dynamic environments.

\begin{figure}[t]
\centering
\includegraphics[width=\columnwidth]{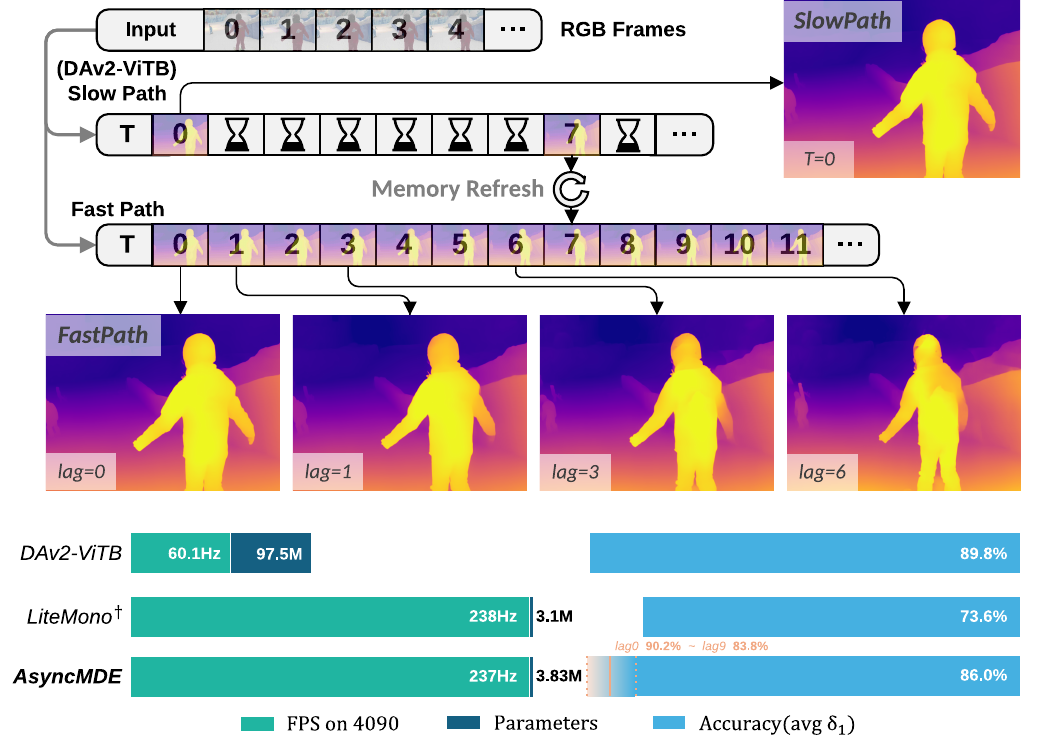}
\caption{Overview of AsyncMDE. \textbf{Top}: the Slow Path (DAv2-ViTB) periodically refreshes spatial memory; the Fast Path fuses cached memory with each frame at high frequency, with depth maps at increasing lag showing bounded degradation. \textbf{Bottom}: efficiency--accuracy trade-off (three-benchmark average $\delta_1$); AsyncMDE (3.83\,M trainable fast path, 237\,FPS fast-path throughput) recovers 77\% of the $\delta_1$ gap between the lightweight baseline and the foundation model.}
\label{fig:teaser}
\end{figure}

\ifcameraready\else
  \vspace*{0.5\baselineskip}
\fi

Beyond single-frame depth estimation, the MDE field continues to advance along two directions. Video depth methods~\cite{shao_learning_2025, hu_depthcrafter_2025, chen_video_2025, chou_flashdepth_2025, kuang_buffer_2025} improve temporal consistency through cross-frame modeling, while Depth Anything~V3~\cite{lin_depth_2025} and VGGT~\cite{wang_vggt_2025} unify monocular and multi-view geometric constraints toward general-purpose visual geometry foundation models. However, embodied intelligence requires not just peak single-frame accuracy, but reliable depth estimation under strict latency and resource constraints---all these methods depend on heavy backbone networks, and real-time edge deployment remains out of reach. Knowledge distillation and lightweight architecture design~\cite{wofk_fastdepth_2019, zhang_lite-mono_2023} attempt to bridge the efficiency gap by compressing model size, but cross-domain generalization and accuracy both drop significantly once the parameter count is reduced to a few million.

The accuracy--efficiency tension extends to the decision-making layer. VLA models such as OpenVLA~\cite{kim_openvla_2025} incur hundreds of milliseconds per inference, far too slow for 50--100\,Hz robotic control loops. Inspired by dual-process theory~\cite{kahneman_thinking_2011}, fast--slow architectures~\cite{han_dual_2024, nvidia_gr00t_2025, cui_openhelix_2025, chen_fast--slow_2025} decouple a slow, large-model system for semantic reasoning from a fast system for high-frequency action execution. This paradigm suggests that a similar fast--slow separation can be applied at the perception level, rather than solely compressing the model itself.

In continuous robot operation, adjacent viewpoints share substantial 3D structure, enabling a decomposition into two subproblems of vastly different complexity: \emph{scene representation}---recovering 3D-aware features from a single 2D image---demands large capacity and strong priors; \emph{temporal adaptation}---incrementally updating cached features for the current viewpoint---is far simpler, as physical continuity bounds inter-frame changes. This gap motivates pairing a heavyweight model that runs infrequently with a lightweight model at high frame rates. We thus propose \textbf{AsyncMDE}, an asynchronous depth perception system that \emph{amortizes} the foundation model's cost over time. The foundation model~\cite{yang_depth_2024} solves scene representation in the background, writing high-quality features into spatial memory; the lightweight network solves temporal adaptation in the foreground, detecting changes and selectively updating memory through complementary fusion. The two paths run concurrently on separate CUDA streams; since the lightweight network only injects changes rather than infers from scratch, it far outperforms distilled models of comparable size. The main contributions are:

\begin{itemize}

\item We propose the asynchronous depth perception paradigm, which exploits the complexity gap between scene representation and temporal adaptation to amortize the foundation model's cost over time. The resulting \emph{rate-controlled perception system}'s accuracy is governed by the hardware-determined refresh rate and scales smoothly with platform capability without retraining.

\item We design SpatialMemoryUnit, which uses complementary fusion and autoregressive memory updates to leverage foundation model features, maintaining bounded accuracy degradation within refresh intervals.

\item The deployed system retains a 97.5M frozen slow path for periodic refresh, while its 3.83M trainable fast path runs at 237~FPS on an RTX~4090 and 161~FPS on a TensorRT-optimized Jetson AGX Orin. We validate AsyncMDE on indoor static, dynamic, and synthetic extreme-motion benchmarks.
\end{itemize}

\section{RELATED WORK}

\subsection{Advances in Monocular Depth Estimation}

Single-frame depth estimation foundation models, including Depth Anything~V2~\cite{yang_depth_2024}, Metric3D~\cite{hu_metric3d_2024}, UniDepthV2~\cite{piccinelli_unidepthv2_2026}, and Depth Pro~\cite{bochkovskiy_depth_2024}, have achieved breakthroughs in cross-domain zero-shot generalization through large-scale pretraining and ViT architectures~\cite{xu_towards_2026}. More recently, Depth Anything~V3~\cite{lin_depth_2025} and VGGT~\cite{wang_vggt_2025} further unify monocular and multi-view geometric constraints within a single framework, moving toward general-purpose visual geometry foundation models. While these efforts continue to raise the accuracy ceiling, they all rely on heavy backbones and high-resolution inference, limiting real-time edge deployment.

A related direction is prior-assisted depth estimation, which fuses RGB with sparse depth or camera parameters~\cite{lin_prompting_2025, liu_manipulation_2025}. Although such methods improve metric accuracy, their additional priors introduce non-negligible overhead and do not address inference efficiency on edge platforms.

Video depth methods~\cite{shao_learning_2025, hu_depthcrafter_2025, chen_video_2025, chou_flashdepth_2025, kuang_buffer_2025} improve temporal consistency through cross-frame modeling. DepthCrafter~\cite{hu_depthcrafter_2025} generates long-sequence consistent depth via video diffusion, and Video Depth Anything~\cite{chen_video_2025} introduces temporal attention heads on top of a single-frame foundation model for inter-frame alignment. These methods improve cross-frame depth quality but still rely on heavy backbones or diffusion sampling processes, making real-time deployment costly. For persistent 3D perception, Spann3R~\cite{wang_3d_2025} and CUT3R~\cite{wang_continuous_2025} use external spatial memory for streaming 3D reconstruction. Their ``persistent-state-driven inference'' paradigm is conceptually close to ours, but they target general geometry estimation and depend on heavy ViT architectures, making them unsuitable for lightweight real-time depth inference.

\subsection{Efficient Depth Perception and Dual-System Architectures}

For depth model compression, knowledge distillation~\cite{wofk_fastdepth_2019, zhang_lite-mono_2023} transfers depth knowledge from large models to lightweight networks via the teacher--student paradigm. However, conventional distillation applies supervision only at the final output level, and the compressed model struggles to inherit the foundation model's rich intermediate representations. In practice, accuracy degradation remains significant when the parameter count is reduced to a few million, indicating that reducing capacity alone cannot preserve both perception quality and real-time performance.

For dual-system architectures, Kahneman's dual-process theory~\cite{kahneman_thinking_2011} provides a cognitive science foundation for fast--slow system separation. Dual MLLM~\cite{han_dual_2024}, GR00T~\cite{nvidia_gr00t_2025}, OpenHelix~\cite{cui_openhelix_2025}, and FiS-VLA~\cite{chen_fast--slow_2025} have validated the effectiveness of this paradigm for robotic decision-making and manipulation. A related systems pattern also appears in visual SLAM, where a fast front end tracks incoming frames while slower back-end optimization refines the global state. This work extends the idea to dense depth perception, using the foundation model as a low-frequency quality source, the lightweight network as a high-frequency executor, and spatial memory as the bridge enabling feature reuse between them.


\begin{table}[t]
\caption{\textsc{Nomenclature}}
\label{tab:notation}
\centering
\renewcommand{\arraystretch}{1.1}
\footnotesize
\begin{tabular}{@{} c l @{}}
\toprule
\textbf{Symbol} & \textbf{Description} \\
\midrule
$x_t \in \mathbb{R}^{3 \times H \times W}$ & Input RGB frame at time $t$ \\
$\hat{y}_t \in \mathbb{R}^{1 \times H \times W}$ & Predicted depth map at time $t$ \\
$N \in \mathbb{Z}^+$ & Refresh interval in frames (for training/evaluation) \\
$f_\theta$ & Lightweight inference network with parameters $\theta$ \\
$F_{\mathrm{B},t}^{(\ell)}$ & Foundation model layer-$\ell$ feature at time $t$ \\
$F_S^{(\ell)}$ & Encoder layer-$\ell$ output feature \\
$\Phi^{(\ell)}$ & Layer-$\ell$ feature projector (Conv$_{1\times1}$+Interp) \\
$T_t^{(\ell)} \in (0,1)$ & Layer-$\ell$ raw spatial modulation factor \\
$T_t^{\prime(\ell)} \in (0,1)$ & Smoothed modulation factor used for fusion \\
$M_t^{(\ell)} \in \mathbb{R}^{128 \times H_\ell \times W_\ell}$ & Layer-$\ell$ spatial memory ($\ell=1,\dots,4$) \\
\bottomrule
\end{tabular}
\end{table}

\section{METHODOLOGY}

This section describes AsyncMDE. Section~III-A outlines the two-phase procedure; III-B describes network components; III-C details spatial memory fusion; and III-D presents the loss. Table~\ref{tab:notation} summarizes the nomenclature.

\subsection{Asynchronous Perception Framework}

\begin{figure*}[t]
\centering
\includegraphics[width=\textwidth]{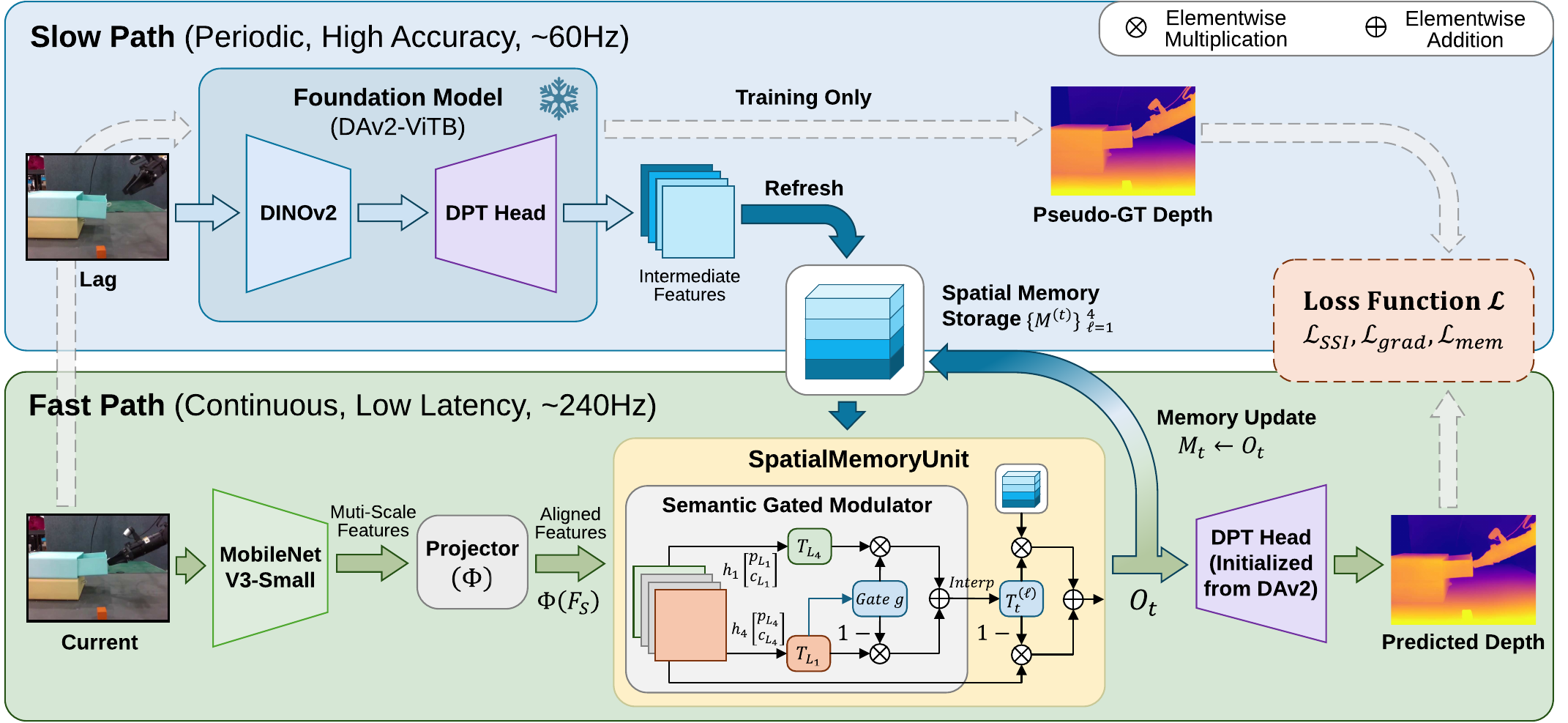}
\caption{AsyncMDE system overview. DAv2-ViTB runs asynchronously in the background (slow path, $\sim$60\,Hz), writing results to spatial memory when available; the lightweight network continuously predicts depth for the current viewpoint (fast path, $\sim$240\,Hz), combining cached memory with current observations through complementary fusion and autoregressively updating memory. During training, DAv2 also provides pseudo-label depth for supervision.}
\label{fig:overview}
\end{figure*}

As shown in Fig.~\ref{fig:overview}, the system maintains a set of multi-scale spatial memories $\{M_t^{(\ell)}\}_{\ell=1}^{4}$, where $\ell$ denotes the feature level, to cache high-quality features from the foundation model and update them autoregressively across frames. The system operates in two phases.

\textbf{Initialization and refresh phase}: The frozen foundation model (DAv2-ViTB~\cite{yang_depth_2024}) initializes memory at $t=0$ and overwrites it whenever a slow-path refresh completes. Let $\mathcal{R}$ denote refresh frames; the memory available to the fast path is
\begin{equation}
M_t^{(\ell)} =
\begin{cases}
F_{\mathrm{B},t}^{(\ell)}, & t=0\ \text{or}\ t \in \mathcal{R}, \\
O_{t-1}^{(\ell)}, & t \notin \mathcal{R}.
\end{cases}
\label{eq:refresh_update}
\end{equation}

\textbf{Continuous inference phase} ($t>0$): The lightweight network runs frame by frame, performing continuous inference through spatial memory fusion:
\begin{equation}
(\hat{y}_t, \, O_t) = f_\theta(x_t, \, M_t)
\end{equation}
where $f_\theta$ consists of an encoder, SpatialMemoryUnit, and decoder; $O_t$ becomes the next memory unless a new slow-path refresh overwrites it. During training, sequence slices with a fixed refresh interval $N$ are used; the foundation model generates pseudo-label depth and refreshes memory, and its parameters are not updated. The foundation model sets the representation quality ceiling, while the lightweight network learns to preserve and adapt it. Implementation and deployment details are provided in Section~IV-B.

\subsection{Network Architecture}

The overall architecture follows three design principles. \emph{Maximize reuse}---the decoder inherits the RefineNet architecture and pretrained weights from the foundation model's DPT Head; the encoder extracts lightweight multi-scale observations; the projector aligns dimensions and SpatialMemoryUnit fuses memory with current features. \emph{Structural minimalism}---no optical flow, depth warping, or attention modules are introduced; all cross-frame information transfer uses per-pixel gated fusion, with only 3.83M trainable parameters (decoder 2.52M, encoder 0.93M, SMU 0.38M). \emph{External state memory}---the update can be interpreted as a gated, spatially varying EMA over explicit memory, but unlike learned recurrent cells, the memory is an externally refreshable spatial cache: $M_0$ is injected by the foundation model, each refresh overwrites memory, and any frame can perform independent inference given memory.

\subsubsection{Encoder}

The encoder employs MobileNetV3-Small~\cite{howard_searching_2019} (0.93M parameters), producing multi-scale features $\{F_S^{(\ell)}\}_{\ell=1}^{4}$ at $4\times$--$32\times$ downsampling. Its role is to provide current-frame observations to SpatialMemoryUnit, not to perform depth estimation independently. Ablation experiments (Section~IV-D) compare multiple encoder architectures, showing that spatial memory compensates for limited encoder capacity.

\subsubsection{Feature Projector}

The projector applies a $1 \times 1$ convolution for channel alignment followed by bilinear interpolation for spatial alignment to each feature level, making encoder outputs dimensionally consistent with the memory features:
\begin{equation}
\Phi^{(\ell)}(F_S^{(\ell)}) = \text{Interp}\big(\text{Conv}_{1\times1}(F_S^{(\ell)}),\; s_\ell\big)
\end{equation}
where $s_\ell$ is the spatial size of the layer-$\ell$ memory. After projection, all levels are unified to 128 channels.

\subsubsection{Decoder}

The decoder directly inherits the RefineNet architecture and pretrained weights from the DPT Head. It progressively fuses the four feature levels from deepest to shallowest, performs cross-scale aggregation through residual convolutional units, and outputs a depth map at the same resolution as the input: $\hat{y}_t \in \mathbb{R}^{1 \times H \times W}$, omitting the batch dimension as in Table~\ref{tab:notation}.

\subsection{Spatial Memory Fusion}

SpatialMemoryUnit is the core component of the system. Viewing the memory update as a discrete-time dynamical system, the fusion form must satisfy boundedness and controllable decay. The per-pixel convex combination $O = T \cdot M + (1-T) \cdot F$ ($T \in (0,1)$) keeps fused features between memory and current observations, preventing long-sequence divergence. When combined with the autoregressive update $M_{t+1}=O_t$, the contribution weight of $M_0$ to frame $t$ becomes $\prod_{s=1}^{t} T_s$, which decays at a rate determined by scene dynamics. This gives a predictable refresh-interval trade-off, verified in Section~IV-C.

\textbf{Justification for feature-space fusion.} Fusion operates in the $8\times$--$32\times$ downsampled feature space, where each vector's receptive field spans hundreds of pixels and encodes semantic and structural information robust to pixel-level displacements. When motion exceeds the receptive field tolerance, $T \to 0$ injects new observations without requiring pose estimation or feature warping, trading a predictable degradation rate for robustness against the known failure modes of optical flow, such as occlusion boundaries and dynamic object interference.

Based on the above, the fusion procedure has two steps. A semantic gated modulation factor assesses regional change (Section~III-C-1), and complementary fusion selectively updates memory (Section~III-C-2).

\subsubsection{Semantic Gated Modulation Factor}

The spatial modulation factor $T_t^{(\ell)} \in (0,1)$ governs the per-pixel trust balance between memory and the current observation. When $T \to 1$ the system retains memory (static region); when $T \to 0$ it injects the current frame (changed region). Shallow features (Layer~1) capture fine-grained texture changes but miss semantic shifts, whereas deep features (Layer~4) capture semantic changes but lack spatial precision. The SemanticGatedModulator fuses both scales through a learnable gating mechanism.

Given the projected features at Layer~1 and 4 from the previous and current frames, the $T$ values at both scales are:
\begin{equation}
T_{L1} = \sigma\big(h^{(1)}([\mathbf{p}_{L1};\, \mathbf{c}_{L1}])\big)
\end{equation}
\begin{equation}
T_{L4} = \sigma\big(h^{(4)}([\mathbf{p}_{L4};\, \mathbf{c}_{L4}])\big)
\end{equation}
where $\mathbf{p}_{Li}$ and $\mathbf{c}_{Li}$ are the projected features of the previous and current frames at Layer~$i$ ($i \in \{1,4\}$), $[\cdot;\cdot]$ denotes channel-wise concatenation, and $h^{(1)}$ and $h^{(4)}$ denote independently parameterized lightweight convolutional networks. The $T$ values from the two scales are non-linearly combined via semantic gating:
\begin{equation}\label{eq:gate}
g = \sigma\big(k \cdot (\widetilde{T}_{L4} - 0.5)\big)
\end{equation}
\begin{equation}\label{eq:tfinal}
T_{\text{final}} = g \odot T_{L1} + (1 - g) \odot \widetilde{T}_{L4}
\end{equation}
where $\widetilde{T}_{L4} = \text{Interp}(T_{L4}, s_1)$ is $T_{L4}$ upsampled to Layer~1 resolution, and $k=4.0$ is a temperature parameter. The gate $g$ causes semantically stable regions to adopt the fine-grained $T_{L1}$, while semantically changing regions switch to $\widetilde{T}_{L4}$ to drive global refresh. The modulation factor for each level is obtained by interpolating $T_{\text{final}}$:
\begin{equation}\label{eq:t_distribute}
T_t^{(\ell)} = \text{Interp}(T_{\text{final}},\; s_\ell), \quad \ell = 1,\dots,4
\end{equation}
where $s_\ell$ is the spatial size of layer $\ell$, and $T_t^{(1)} = T_{\text{final}}$.

Additionally, the system applies temporal smoothing $T_t^{\prime(\ell)} = \beta T_t^{(\ell)} + (1-\beta) T_{t-1}^{\prime(\ell)}$ with $\beta=0.5$, suppressing frame-to-frame jitter in the $T$ values.

\subsubsection{Complementary Fusion and Memory Update}

Given the smoothed spatial modulation factor $T_t^{\prime(\ell)}$, SpatialMemoryUnit mixes memory and current observations per pixel through complementary fusion:
\begin{equation}\label{eq:fusion}
O_t^{(\ell)} = T_t^{\prime(\ell)} \odot M_t^{(\ell)} + (1 - T_t^{\prime(\ell)}) \odot \Phi^{(\ell)}(F_{S,t}^{(\ell)})
\end{equation}
The fused result is written back as an autoregressive memory update, sustaining foundation-model feature quality within the refresh interval:
\begin{equation}\label{eq:memory_update}
M_{t+1}^{(\ell)} = O_t^{(\ell)}
\end{equation}
where $M_t^{(\ell)} \in \mathbb{R}^{B \times 128 \times H_\ell \times W_\ell}$ is the layer-$\ell$ spatial memory at time $t$, and $\Phi^{(\ell)}$ is the feature projector.

\subsection{Loss Function}

The training objective consists of three loss terms that constrain depth accuracy, edge structure, and memory utilization, respectively.

\textbf{Scale-Shift Invariant Loss} eliminates global scale and shift differences between the predicted depth $P$ and the pseudo-label $G$:
\begin{equation}
\mathcal{L}_{\text{SSI}} = \text{MSE}\!\left(\frac{P - \mu_P}{\sigma_P},\;\frac{G - \mu_G}{\sigma_G}\right)
\end{equation}
where $\mu$ and $\sigma$ denote the spatial mean and standard deviation, respectively.

\textbf{Multi-Scale Gradient Loss} enforces edge sharpness on the normalized depth maps:
\begin{equation}
\mathcal{L}_{\text{grad}} = \sum_{s=1}^{4} \frac{1}{s^2}\big(|\nabla_x P_s - \nabla_x G_s| + |\nabla_y P_s - \nabla_y G_s|\big)
\end{equation}
where $s$ indicates the downsampling scale; gradients are computed at multiple scales to capture both fine and coarse edge structure.

\textbf{Memory Regularization Loss} prevents the network from ignoring the memory and relying solely on the current frame during early training by introducing a soft lower bound on the $T$ values:
\begin{equation}
\mathcal{L}_{\text{mem}} = \text{ReLU}(\tau - \bar{T})
\end{equation}
where $\bar{T}$ is the spatial mean of the Layer~1 $T$ values and $\tau = 0.4$ is the lower-bound threshold. This constraint activates only when $\bar{T}$ falls below the threshold, without interfering with the network's freedom to assign low $T$ values to local dynamic regions.

The total loss is:
\begin{equation}
\mathcal{L} = 1.0 \cdot \mathcal{L}_{\text{SSI}} + 0.5 \cdot \mathcal{L}_{\text{grad}} + 0.1 \cdot \mathcal{L}_{\text{mem}}
\end{equation}


\begin{table*}[!t]
\caption{\textsc{Accuracy--Efficiency Comparison}}
\label{tab:main_results}
\centering
\renewcommand{\arraystretch}{1.1}
\footnotesize
\begin{tabular}{@{} l c c c c c c c c c c c @{}}
\toprule
& & & \multicolumn{3}{c}{\textbf{ScanNet}} & \multicolumn{3}{c}{\textbf{Bonn}} & \multicolumn{3}{c}{\textbf{Sintel}} \\
\cmidrule(lr){4-6} \cmidrule(lr){7-9} \cmidrule(lr){10-12}
\textbf{Method} & \textbf{Params} & \textbf{FPS} & \textbf{AbsRel$\downarrow$} & \textbf{RMSE$\downarrow$} & \textbf{$\delta_1$$\uparrow$} & \textbf{AbsRel$\downarrow$} & \textbf{RMSE$\downarrow$} & \textbf{$\delta_1$$\uparrow$} & \textbf{AbsRel$\downarrow$} & \textbf{RMSE$\downarrow$} & \textbf{$\delta_1$$\uparrow$} \\
\midrule
DAv2-ViTL~\cite{yang_depth_2024} & 335.3M & 21.1 & \textbf{0.037} & \textbf{0.133} & \textbf{0.984} & 0.049 & 0.174 & 0.979 & 0.224 & 4.985 & 0.735 \\
DAv2-ViTB & 97.5M & 60.1 & 0.040 & 0.137 & 0.983 & 0.050 & 0.177 & 0.979 & 0.222 & \textbf{4.867} & 0.733 \\
DAv2-ViTS & 24.8M & 132.1 & 0.044 & 0.146 & 0.980 & 0.052 & 0.178 & 0.979 & 0.235 & 4.993 & 0.717 \\
VDA-Base~\cite{chen_video_2025} & 114.4M & 42.4 & 0.041 & 0.144 & 0.982 & \textbf{0.042} & 0.170 & 0.985 & \textbf{0.212} & 4.922 & \textbf{0.758} \\
VDA-Small & 29.0M & 75.9 & 0.047 & 0.154 & 0.979 & 0.044 & \textbf{0.160} & \textbf{0.986} & 0.233 & 5.040 & 0.716 \\
CUT3R~\cite{wang_continuous_2025} & 748.4M & 14.4 & 0.107 & 0.390 & 0.888 & 0.112 & 0.377 & 0.877 & 0.448 & 6.185 & 0.425 \\
LiteMono~\cite{zhang_lite-mono_2023} & \textbf{3.07M} & \textbf{238} & 0.168 & 0.431 & 0.727 & 0.158 & 0.455 & 0.759 & 0.416 & 5.956 & 0.443 \\
LiteMono$^\dagger$ & \textbf{3.07M} & \textbf{238} & 0.120 & 0.306 & 0.851 & 0.123 & 0.365 & 0.854 & 0.383 & 5.664 & 0.502 \\
\midrule
AsyncMDE (ours) & 3.83M(+97.5M) & 237 & 0.057 & 0.181 & 0.968 & 0.058 & 0.196 & 0.969 & 0.287 & 5.377 & 0.640 \\
\bottomrule
\end{tabular}
\par\vspace{4pt}
\raggedright\scriptsize
All baselines perform independent per-frame inference; AsyncMDE reports lag\,0--9 cycle averages. For AsyncMDE, Params denote trainable fast path plus frozen slow path, and FPS denotes fast-path throughput. $^\dagger$\,Fine-tuned on our training data using the same frozen-foundation pseudo-label supervision as AsyncMDE. LiteMono input $640{\times}192$, CUT3R input $224{\times}224$; all others are resized so that the shorter side equals 518\,px (divisible by 14).
\end{table*}
\section{EXPERIMENTS}\label{sec:exp}

\subsection{Experimental Setup}

\subsubsection{Training Setup}

Training mixes three datasets: NYUv2~\cite{silberman_indoor_2012} (indoor static, $\sim$155K frames), TartanAir~\cite{wang_tartanair_2020} (synthetic, $\sim$112K frames), and BridgeData~V2~\cite{walke_bridgedata_2023} (robotic interaction, $\sim$387K frames), totaling $\sim$654K frames. The supervision signal is pseudo-labeled inverse depth generated by the frozen foundation model. Training uses video clips of length 10 with stride 5; each image is resized so that its short edge measures 518 pixels, aligned to multiples of 14. We use AdamW ($\text{lr}{=}10^{-4}$, weight decay 0.01), batch size 4, 30 epochs, and cosine learning rate scheduling. AMP is enabled with gradient clipping at 1.0.

\subsubsection{Evaluation Setup}

Evaluation is conducted on three benchmarks covering different scene characteristics. \textbf{ScanNet}~\cite{dai_scannet_2017} contains 100 indoor scenes (43{,}600 frames, structured-light depth GT). \textbf{Bonn}~\cite{palazzolo_refusion_2019} contains 14 dynamic indoor scenes (2{,}850 frames). \textbf{Sintel}~\cite{butler_naturalistic_2012} contains 23 synthetic sequences (1{,}064 frames) with large motion and extreme dynamics. The evaluation refresh interval is $N{=}10$. Depth accuracy metrics are AbsRel, RMSE, and $\delta_1$ (the fraction of pixels satisfying $\max(d^*/d, d/d^*) < 1.25$); efficiency metrics report fast-path FPS and distinguish trainable fast-path parameters from frozen slow-path parameters for AsyncMDE.

\textbf{Metric reporting convention.} The accuracy of AsyncMDE varies with lag (the number of frames since the last refresh). All values reported in tables are averages over one complete refresh cycle ($\text{lag}\in[0,N{-}1]$), referred to as the \emph{cycle average}; per-lag degradation behavior is presented in Section~IV-C.

\subsubsection{Baselines}

DAv2-ViTB (97.5M) serves as the primary baseline and is also the frozen foundation model used for the slow path; DAv2-ViTS/ViTL serve as same-family references. The baselines also include LiteMono~\cite{zhang_lite-mono_2023} (3.07M, as a lightweight representative), Video Depth Anything~\cite{chen_video_2025} (as a video depth representative), and CUT3R~\cite{wang_continuous_2025} (streaming 3D reconstruction with external spatial memory). Because LiteMono performed poorly with its original weights due to limited training data, we fine-tuned it on our dataset using the same frozen-foundation pseudo-label supervision as AsyncMDE, denoted LiteMono$^\dagger$.

\subsection{Main Results: Accuracy--Efficiency Comparison}

\textbf{Accuracy comparison with baselines.} As shown in Table~\ref{tab:main_results}, AsyncMDE achieves $\delta_1$=96.8\% and 96.9\% on ScanNet and Bonn, respectively, using a 3.83M-parameter trainable fast path together with a 97.5M frozen slow path, keeping the gap within 2 percentage points of DAv2-ViTB (98.3\%/97.9\%). This comparison highlights the value of spatial memory from three perspectives. (1)~\emph{vs.\ standalone lightweight models.} Compared with LiteMono$^\dagger$ at a similar trainable fast-path scale, AbsRel is 52\% lower, demonstrating that feature amortization is far superior to standalone lightweight models of comparable size. (2)~\emph{vs.\ general-purpose memory architectures.} CUT3R also uses external memory but targets general 3D reconstruction, achieving only 88.8\% $\delta_1$ with 748M parameters, still inferior to AsyncMDE, confirming that task-specific design is essential for efficient amortization. (3)~\emph{vs.\ extreme scenarios.} On Sintel, the cycle-average AbsRel of 0.287 reflects the inherent limitation of the temporal continuity prior under severe scene dynamics; however, the degradation is strictly lower-bounded by the standalone encoder capacity (AbsRel=0.386), confirming that the system retains a guaranteed performance floor.

\begin{figure}[t]
\centering
\includegraphics[width=\columnwidth]{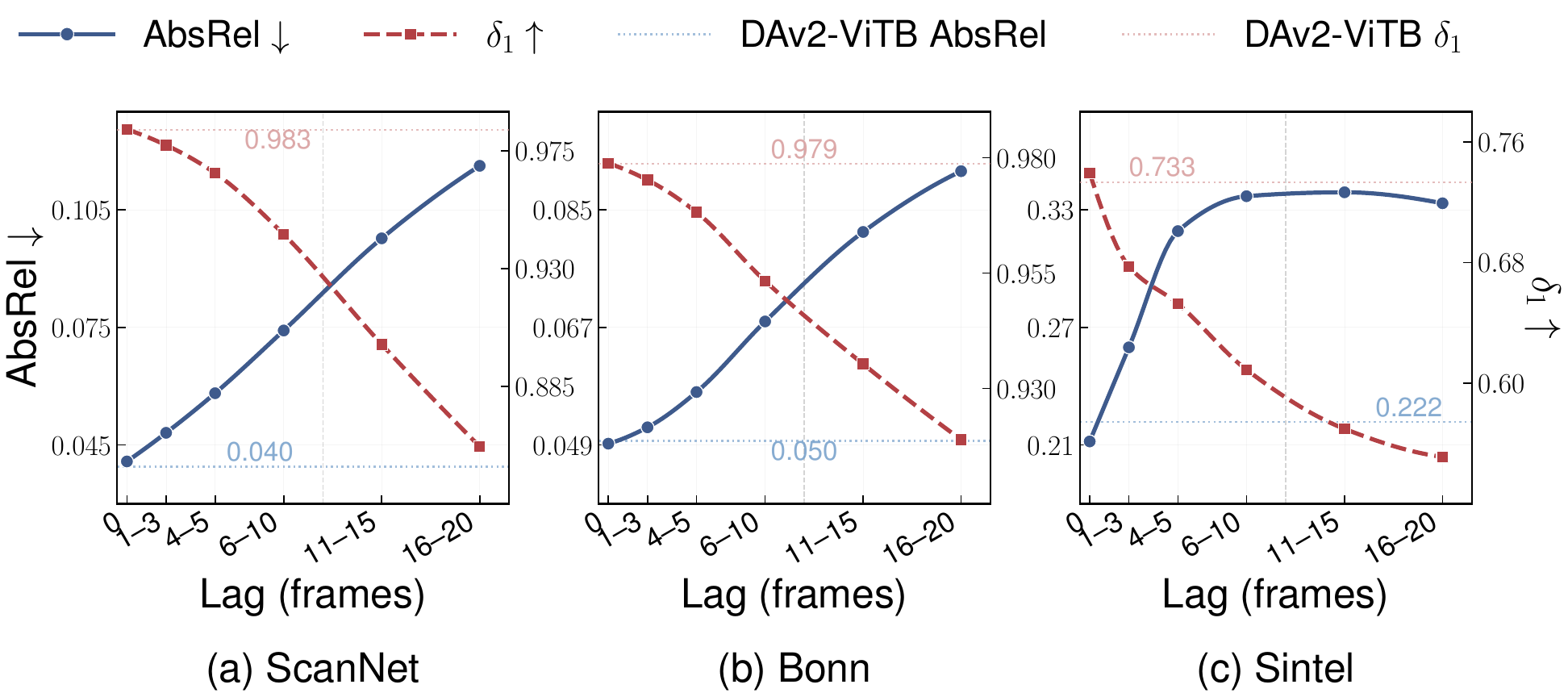}
\caption{Lag--accuracy degradation curves. The evaluation interval $N{=}20$ exceeds the training setting ($N{=}10$) to test out-of-distribution generalization. ScanNet and Bonn degrade predictably within the training interval (lag$\leq$10) and more steeply beyond; Sintel AbsRel saturates beyond lag${=}$10 at $\sim$0.34, exhibiting bounded degradation.}
\label{fig:lag_curve}
\end{figure}

\textbf{Accuracy--latency coupling and deployment architecture.} Unlike conventional models with deterministic accuracy, AsyncMDE's accuracy is governed by the rate ratio of the fast and slow paths. At deployment, the two paths occupy separate CUDA streams and exchange data through a lock-free shared feature cache. The slow path writes updated multi-scale features into the cache whenever a new foundation-model inference completes, and the fast path reads the latest cached features at each frame without blocking. The fast-path latency is $t_{\text{fast}}{=}4.2$\,ms; the slow-path latency $t_{\text{slow}}{=}16.6$\,ms is entirely hidden by the pipeline, yielding an effective refresh interval $N_{\text{eff}} \approx 237/60 \approx 4$ frames on an RTX~4090. As platform compute capability varies, $N_{\text{eff}}$ adjusts accordingly and accuracy traces the degradation curve smoothly (Fig.~\ref{fig:lag_curve}). Section~IV-C quantifies this property.

\subsection{System Behavior Analysis: Degradation and Deployment}

\subsubsection{Accuracy Degradation Characteristics}

Fig.~\ref{fig:lag_curve} shows the relationship between lag and accuracy, with $N$ set to 20 at evaluation to test generalization beyond the training distribution. At lag=0, ScanNet AbsRel is 0.041, nearly identical to DAv2-ViTB (0.040). As lag increases, degradation exhibits two key properties. (1)~\textbf{Predictable and bounded.} ScanNet and Bonn degrade gradually within the training interval (lag$\leq$10), consistent with the theory in Section~III-C; Sintel saturates for lag$>$10 at $\sim$0.34, with the lower bound being the encoder's standalone inference capability (FastPath Only AbsRel=0.386). (2)~\textbf{Scene-dependent.} Bonn has a small fraction of dynamic objects, with AbsRel increasing only 38\% relative to lag=0 at lag\,6--10; ScanNet increases by 82\% due to global camera scanning; Sintel exhibits the steepest rise (AbsRel reaches 0.337) yet saturates beyond lag$>$10, indicating bounded degradation even under extreme dynamics.

\subsubsection{Amortization Efficiency and Deployment Trade-offs}

When $N$ increases to 20, the ScanNet cycle-average AbsRel rises from 0.057 to 0.081 ($+$42\%), and $\delta_1$ drops from 96.8\% to 92.6\%; the degradation magnitude can be estimated from Fig.~\ref{fig:lag_curve}. On edge platforms, the slow-path rate decreases and $N_{\text{eff}}$ increases, but the fast-path frame rate is unaffected. To verify edge feasibility, we conduct inference benchmarks on a Jetson AGX Orin (64\,GB, 50\,W mode).

\begin{table}[b]
\caption{\textsc{Jetson AGX Orin Edge Deployment, Input $518{\times}518$}}
\label{tab:jetson}
\centering
\renewcommand{\arraystretch}{1.1}
\setlength{\tabcolsep}{3pt}
\footnotesize
\begin{tabular}{@{}l ccc ccc@{}}
\toprule
& \multicolumn{3}{c}{\textbf{PyTorch FP32}} & \multicolumn{3}{c}{\textbf{TRT FP16}} \\
\cmidrule(lr){2-4} \cmidrule(lr){5-7}
& \textbf{FastPath} & \textbf{ViTS} & \textbf{ViTB} & \textbf{FastPath} & \textbf{ViTS} & \textbf{ViTB} \\
\midrule
FPS$\uparrow$ & 27.8 & 10.1 & 4.1 & \textbf{161.1} & 25.3 & 12.3 \\
Latency\,(ms) & 36.0 & 99.3 & 244.0 & \textbf{6.2} & 39.5 & 81.0 \\
\midrule
Speedup\,(vs ViTB) & \multicolumn{3}{c}{6.8$\times$} & \multicolumn{3}{c}{\textbf{13.1$\times$}} \\
\bottomrule
\end{tabular}
\end{table}

Table~\ref{tab:jetson} presents the Orin edge deployment characteristics. Under PyTorch, the fast path runs at 27.8\,FPS, a 6.8$\times$ speedup over DAv2-ViTB (higher than the 3.9$\times$ on RTX~4090), as depthwise separable convolutions exhibit a more pronounced efficiency advantage on edge devices; the SMU module accounts for only 14\% of latency (5.0\,ms), confirming that memory fusion overhead is minimal. TRT FP16 further boosts the fast path to 161.1\,FPS (13.1$\times$), at which point $N_{\text{eff}} \approx 161/12 \approx 13$ frames, still within the bounded degradation region of Fig.~\ref{fig:lag_curve}. Note that under single-GPU concurrency, the fast path drops to 13.5\,FPS (2.16$\times$ slowdown) due to resource contention; TRT deployment or a dual-GPU setup can mitigate this bottleneck.

\begin{table}[!t]
\caption{\textsc{Encoder Ablation on ScanNet, N=10 Cycle Average}}
\label{tab:encoder_ablation}
\centering
\renewcommand{\arraystretch}{1.1}
\setlength{\tabcolsep}{5pt}
\footnotesize
\begin{tabular}{@{} l c c c c c c @{}}
\toprule
\textbf{Encoder} & \textbf{Type} & \textbf{Params} & \textbf{FPS$\uparrow$} & \textbf{AbsRel$\downarrow$} & \textbf{$\delta_1$$\uparrow$} & \textbf{FastPath\%} \\
\midrule
\textbf{MNv3-S} & CNN & \textbf{3.83M} & \textbf{237} & \textbf{0.057} & \textbf{0.968} & 11.3 \\
MNv3-L & CNN & 5.93M & 198 & 0.060 & 0.962 & 12.7 \\
EViT-B0 & Linear & 3.53M & 207 & 0.065 & 0.955 & 23.4 \\
EViT-B1 & Linear & 7.52M & 147 & 0.064 & 0.958 & 37.6 \\
MViT-XS & Hybrid & 4.82M & 112 & 0.070 & 0.945 & 39.7 \\
MViT-S & Hybrid & 7.87M & 110 & 0.065 & 0.956 & 33.7 \\
\bottomrule
\end{tabular}
\end{table}

\begin{table}[b]
\caption{\textsc{Core Architecture Ablation, N=10 Cycle Average}}
\label{tab:ablation_core}
\centering
\renewcommand{\arraystretch}{1.1}
\setlength{\tabcolsep}{2pt}
\footnotesize
\begin{tabular}{@{}l ccc cc cc cc@{}}
\toprule
& & & & \multicolumn{2}{c}{\textbf{ScanNet}} & \multicolumn{2}{c}{\textbf{Bonn}} & \multicolumn{2}{c}{\textbf{Sintel}} \\
\cmidrule(lr){5-6} \cmidrule(lr){7-8} \cmidrule(lr){9-10}
\textbf{Config.} & \textbf{M} & \textbf{Init} & \textbf{SGM} & \textbf{AR$\downarrow$} & \textbf{$\delta_1\!\uparrow$} & \textbf{AR$\downarrow$} & \textbf{$\delta_1\!\uparrow$} & \textbf{AR$\downarrow$} & \textbf{$\delta_1\!\uparrow$} \\
\midrule
FastPath Only & \ding{55} & -- & -- & 0.132 & 0.825 & 0.114 & 0.873 & 0.386 & 0.492 \\
Enc. Mem. & \ding{51} & Enc. & Full & 0.141 & 0.801 & 0.112 & 0.879 & 0.432 & 0.444 \\
DAv2 w/o SGM & \ding{51} & DAv2 & L1 & 0.061 & 0.960 & 0.062 & 0.965 & 0.297 & 0.637 \\
DAv2 L4 Only & \ding{51} & DAv2 & L4 & \textbf{0.057} & 0.966 & \textbf{0.056} & \textbf{0.970} & \textbf{0.287} & 0.637 \\
\textbf{Full SMU} & \ding{51} & DAv2 & Full & \textbf{0.057} & \textbf{0.968} & 0.058 & 0.969 & \textbf{0.287} & \textbf{0.640} \\
\bottomrule
\end{tabular}
\end{table}

\subsection{Ablation Studies}

\subsubsection{Encoder Selection Ablation}

Table~\ref{tab:encoder_ablation} fixes SpatialMemoryUnit and the decoder, varying the encoder across pure CNN (MobileNetV3~\cite{howard_searching_2019}), hybrid architectures (MobileViT~\cite{mehta_mobilevit_2021}), and linear attention (EfficientViT~\cite{cai_efficientvit_2023}).

The results show a counterintuitive trend: larger encoder capacity leads to lower system accuracy. MNv3-S achieves the best overall performance; EViT-B1 worsens AbsRel by $+$12\% and reduces FPS by 38\%. This is explained by FastPath\% (the fraction of encoder-dominated regions): MNv3-S has only 11.3\% encoder-dominated regions, with 88.7\% retaining high-quality memory; EViT-B1 (37.6\%) and MViT-XS (39.7\%) excessively overwrite memory with lower-quality observations. This indicates that the optimal role of the encoder is as a \emph{change detector and observation injector}---injecting new observations only where memory has become invalid, rather than performing independent depth inference.

\subsubsection{Core Architecture Ablation}

Table~\ref{tab:ablation_core} progressively introduces core components to quantify each design choice.

Using FastPath Only (no memory) as an anchor, initializing memory with encoder features (Enc.\ Mem.) actually increases ScanNet AbsRel by $+$7\%, as low-quality features introduce drift through autoregressive accumulation. Switching to DAv2 initialization reduces AbsRel by 57\% and improves $\delta_1$ by 14.3 pp, establishing the value of high-quality memory. DAv2 L4 Only (semantic gating only) closely matches Full SMU (ScanNet AbsRel 0.057 vs.\ 0.057), while DAv2 w/o SGM (texture gating only) degrades to 0.061 ($+$7\%), indicating semantic features are the primary gating contributor. Dual-scale fusion (Full SMU) still holds a slight $\delta_1$ edge, as texture information contributes to depth details.

\begin{figure*}[!t]
\centering
\includegraphics[width=\textwidth]{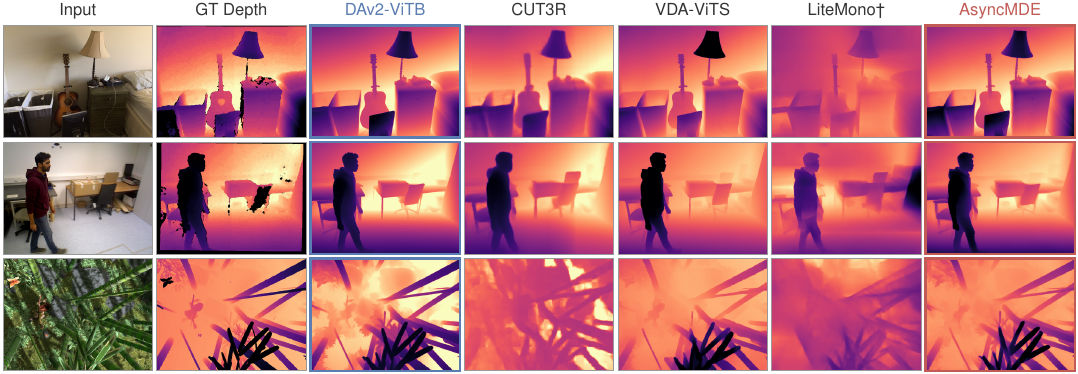}
\caption{Qualitative depth comparison (least-squares aligned). The three rows correspond to ScanNet (indoor static), Bonn (indoor dynamic), and Sintel (synthetic extreme). AsyncMDE produces depth quality comparable to DAv2-ViTB at low lag.}
\label{fig:qualitative}
\vspace{12pt}
\includegraphics[width=\textwidth]{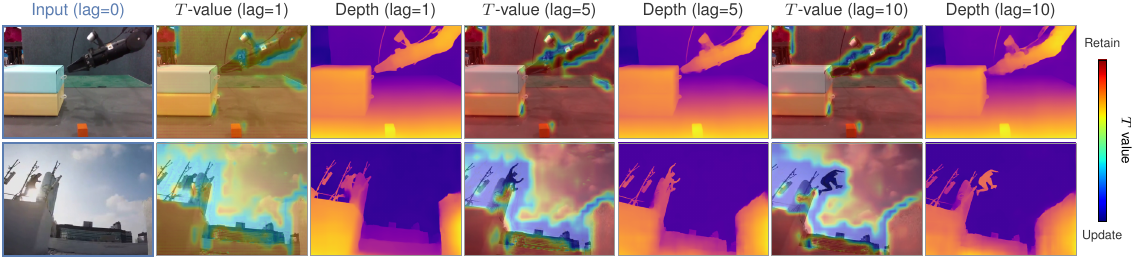}
\caption{$T$-value visualization. Each row shows, from left to right, the refresh-frame RGB, $T$-value masks and depth estimates at lag=1 and lag=$N$. Top: indoor robotic manipulation; bottom: outdoor dynamic scene. Warm colors ($T\!\to\!1$) indicate static regions; cool colors ($T\!\to\!0$) indicate moving regions. Degradation primarily affects moving objects, while static structures maintain stable estimates even at high lag.}
\label{fig:t_value}
\end{figure*}

\subsection{Qualitative Results}

Fig.~\ref{fig:qualitative} presents a qualitative depth comparison. In ScanNet scenes, AsyncMDE's edge sharpness and low-texture smoothness are close to those of DAv2-ViTB and far superior to LiteMono; in Bonn dynamic scenes, moving object contours are well preserved; in Sintel, distant details show some degradation, consistent with the quantitative findings.

Fig.~\ref{fig:t_value} visualizes the spatial evolution of $T$ values across lag. In the indoor robotic scene (top row), $T \approx 1$ everywhere at lag=1 except at the robot arm's end effector; by lag=$N$, the low-$T$ region expands along the motion trajectory with local depth blurring, while static structures retain high $T$. In the outdoor scene (bottom row), the parkour figure triggers low $T$ even at lag=1 and the affected area grows markedly by lag=$N$, whereas the building background stays at $T \to 1$ throughout. Two patterns emerge: (1)~$T$ values distinguish static from dynamic regions, preserving estimate quality for static structures even at high lag; (2)~degradation primarily affects moving objects, consistent with Section~IV-C.

\section{CONCLUSIONS}\label{sec:conclusion}

This paper presents AsyncMDE, which bridges the gap between high-accuracy monocular depth estimation and real-time deployment by amortizing the computational cost of a foundation model over time rather than replacing it with a standalone compressed model. The fast path achieves 237\,FPS (RTX~4090) / 161\,FPS (Orin TRT) with 3.83M trainable parameters, while the deployed system retains a 97.5M frozen slow path for periodic refresh. The system maintains bounded degradation within refresh intervals, and the asynchronous amortization paradigm generalizes to dense perception tasks that rely on spatiotemporal continuity.

\textbf{Limitations and future work.} We identify two limitations that also suggest future research directions. \emph{Extreme motion degradation}---when scene motion causes large-scale memory invalidation (an excessive fraction of $T \to 0$), the system falls back to standalone encoder inference, reaching the inherent lower bound of the temporal continuity prior. Motion-adaptive memory reset or region-specific memory update strategies could address this. \emph{Scale consistency}---the current system outputs relative depth without inter-frame metric scale constraints. For applications that require absolute depth, such as robot navigation, incorporating a temporal scale alignment module or integrating with metric depth foundation models would be necessary.

{\footnotesize\linespread{1}\selectfont
\renewcommand{\url}[1]{}%
\bibliographystyle{IEEEtran}
\bibliography{IEEEabrv,ref}
}

%
%

\end{document}